\pdfoutput=1

\documentclass[11pt]{article}

\usepackage[final]{acl}

\usepackage{times}
\usepackage{latexsym}

\usepackage[T1]{fontenc}

\usepackage[utf8]{inputenc}

\usepackage{microtype}

\usepackage{inconsolata}

\usepackage{graphicx}
\usepackage{balance}
\usepackage{graphicx}
\usepackage{subfigure}
\usepackage{caption}
\usepackage{subcaption}
\usepackage{enumitem}
\usepackage{multirow}
\usepackage{pifont}
\usepackage{amsmath}
\usepackage{amssymb}
\usepackage{footnote}

\usepackage{tabularx}

\usepackage{color}
\usepackage{array}
\usepackage{booktabs}
\usepackage{tcolorbox}
\definecolor{customgray}{rgb}{0.5,0.5,0.5}
\title{Enhancing Large Language Model for Knowledge Graph Completion via Structure-Aware Alignment-Tuning}

\author{
 \textbf{Yu Liu\textsuperscript{1,2}},
 \textbf{Yanan Cao\textsuperscript{1,2}},
 \textbf{Xixun Lin\textsuperscript{1,2}}\thanks{ Corresponding author.},
 \textbf{Yanmin Shang\textsuperscript{1,2}},
 \textbf{Shi Wang\textsuperscript{3}},
 \textbf{Shirui Pan\textsuperscript{4}}
\\
 \textsuperscript{1}Institute of Information Engineering, Chinese Academy of Sciences, Beijing, China
\\
 \textsuperscript{2}School of Cyber Security, University of Chinese Academy of Sciences, Beijing, China
\\
 \textsuperscript{3}Institute of Computing Technology, Chinese Academy of Sciences, Beijing, China
\\
 \textsuperscript{4}Griffith University, Queensland, Australia
\\
\texttt{\{liuyu2022,caoyanan,linxixun,shangyanmin\}@iie.ac.cn}
\\
\texttt{wangshi@ict.ac.cn, s.pan@griffith.edu.au}
}

\begin{document}
\maketitle
\begin{abstract}
Knowledge graph completion (KGC) aims to infer new knowledge and make predictions from knowledge graphs. Recently, large language models (LLMs) have exhibited remarkable reasoning capabilities. LLM-enhanced KGC methods primarily focus on designing task-specific instructions, achieving promising advancements. However, there are still two critical challenges. First, existing methods often ignore the inconsistent representation spaces between natural language and graph structures. Second, most approaches design separate instructions for different KGC tasks, leading to duplicate works and time-consuming processes. To address these challenges, we propose SAT, a novel framework that enhances LLMs for KGC via structure-aware alignment-tuning. Specifically, we first introduce hierarchical knowledge alignment to align graph embeddings with the natural language space through multi-task contrastive learning. Then, we propose structural instruction tuning to guide LLMs in performing structure-aware reasoning over KGs, using a unified graph instruction combined with a lightweight knowledge adapter. Experimental results on two KGC tasks across four benchmark datasets demonstrate that SAT significantly outperforms state-of-the-art methods, especially in the link prediction task with improvements ranging from 8.7\% to 29.8\%\footnote{Our source code is available at \url{https://github.com/liuyudiy/SAT}.}. 

 





\end{abstract}

\begin{figure}
    \centering
    \includegraphics[width=0.49\textwidth]{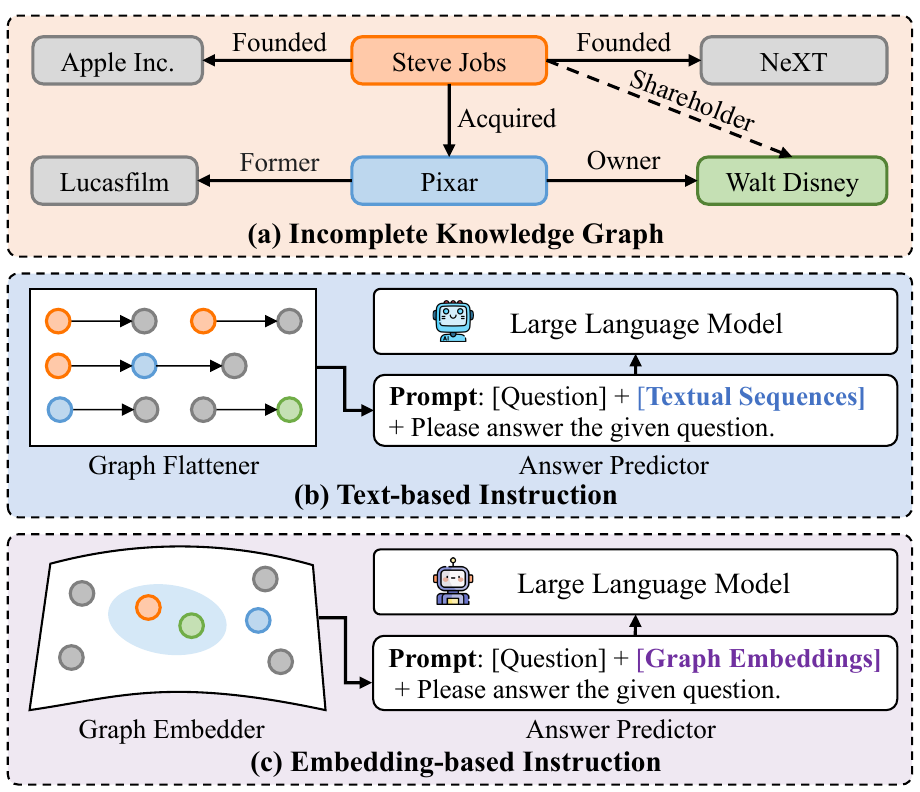}
    \caption{Illustration of LLM-enhanced KGC methods. (a) Incomplete knowledge graph with missing triples. (b) Text-based methods flattening the KG into textual sequences. (c) Embedding-based methods integrating graph embeddings into LLMs.} \label{fig_example}
\end{figure}

\section{Introduction}
Knowledge graphs (KGs) organize world knowledge with structured relations between entities~\citep{bordes2013translating, chaudhri2022knowledge}. In recent years, KGs have gained significant attention in various fields, such as information retrieval~\citep{liu2018entity}, question answering~\citep{luo2024reasoning}, and recommendation systems~\citep{guo2020survey}. Despite great achievements, real-world KGs often suffer from incompleteness, as shown in Figure~\ref{fig_example}(a), which inevitably limits their practical applications~\citep{ li2022does, liu2024generative}. Therefore, it is necessary to develop knowledge graph completion (KGC)~\citep{wang2021kepler} to automatically infer missing triples.






Recently, large language models (LLMs) have demonstrated outstanding performance in various natural language processing (NLP) tasks~\citep{meyer2023chatgpt}. LLM-enhanced KGC aims to leverage the generalizability of LLMs to make predictions over KGs. Existing methods can be broadly categorized into two lines: \textit{text-based} and \textit{embedding-based} methods, as illustrated in Figure~\ref{fig_example}(b) and (c). Text-based methods~\citep{wei2023kicgpt, xu2024multi} verbalize relevant KG triples and append them to input prompts. While these approaches provide explicit paths, the flattening process destroys the underlying structure of KGs. In contrast, embedding-based methods~\citep{ye2023natural, zhang2024making, tang2024graphgpt} employ graph representation learning ~\citep{kipf2016semi,lin2024graph} to generate graph embeddings for the retrieved subgraphs. By preserving the structures within KGs, these methods enable more robust reasoning and achieve impressive performance.






Despite the aforementioned success, embedding-based methods still face two critical challenges. First, while existing methods leverage graph embeddings to capture structural information, they often ignore the representational gap between graph structures and natural language, limiting LLMs’ ability to fully comprehend and reason with structural knowledge.
Second, current research predominantly depends on instruction tuning; however, most studies craft separate instructions for different KGC tasks (e.g., triple classification and link prediction), leading to redundancy and inefficiency. Therefore, it is essential to develop a new instruction-tuning paradigm that integrates various KGC tasks into a unified framework.

To overcome these challenges, we propose a novel framework named \textbf{SAT}, designed to enhance LLMs for KGC through \textbf{S}tructure-aware \textbf{A}lignment and \textbf{T}uning. Specifically, SAT first introduces a \textit{hierarchical knowledge alignment} module to improve LLMs' understanding of graph structure encoding. We construct graph-text alignment datasets from a hierarchical perspective, considering both node-level and subgraph-level alignments, and employ multi-task contrastive learning to align graph embeddings with the natural language space. Subsequently, a \textit{structural instruction tuning} module is proposed to guide LLMs in performing structure-aware reasoning. We design a new graph instruction and implement a lightweight tuning strategy, using a knowledge adapter to accommodate various KGC tasks.

In summary, our contributions are as follows:
\begin{itemize}
\setlength\itemsep{0em}
\item We present a new framework that seamlessly integrates graph structures into LLMs for KGC via structure-aware alignment-tuning.
\item We propose hierarchical knowledge alignment to address the inconsistent representation spaces and structural instruction tuning to efficiently unify diverse KGC tasks.
\item We evaluate SAT on two KGC tasks across four datasets. Extensive experiments show that our model surpasses state-of-the-art methods, particularly in the link prediction task.
\end{itemize}




\begin{figure*}
    \centering
    \includegraphics[width=0.99\textwidth]{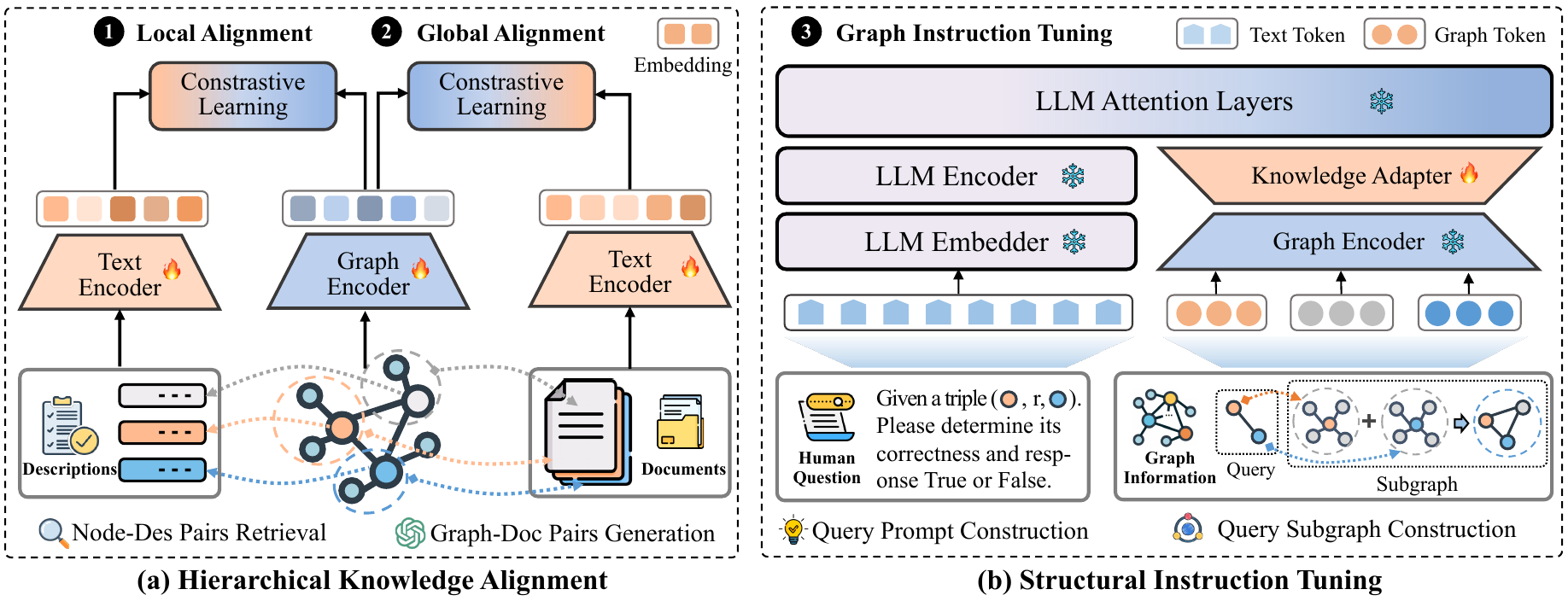}
	\caption{The overall framework of our \textbf{SAT}. (a) Hierarchical knowledge alignment, comprising local and global alignments, aligns graph structural representations with the natural language space. (b) Structural instruction tuning enables LLMs to perform structure-aware reasoning over KGs through a lightweight tuning strategy.} \label{fig_framework}
\end{figure*}

\section{Related work}

Nowadays, LLM-enhanced KGC has garnered increasing interest. Existing works can generally be categorized into two main directions.

Text-based methods verbalize relevant KGs to augment instructions for LLMs. KG-LLM~\citep{yao2023exploring} is the first to explore various LLMs for KGC. KICGPT~\citep{wei2023kicgpt} integrates LLMs with a triple-based retriever, while MPIKGC~\citep{xu2024multi} employs LLMs to generate auxiliary texts for traditional KGC models.  CP-KGC~\citep{yang2024enhancing} and GS-KGC~\citep{yang2024exploiting} further incorporate additional contextual constraints. Moreover, several methods~\citep{jiang2024kgnips, chen2024new} leverage LLMs for the enrichment of KG structures. Despite their success, these approaches still suffer from missing vital triples that make some questions unanswerable. Additionally, the flattening process may undermine the underlying structures within KGs.



Embedding-based methods incorporate structural knowledge into LLMs through graph representation learning. CSProm-KG~\citep{chen2023dipping} and PDKGC~\citep{geng2023prompting} investigate pre-trained language models (PLMs) for KGC. More recently, KoPA~\citep{zhang2024making} has utilized a prefix adapter to fuse structural embeddings into LLMs. MKGL~\citep{guo2024mkgl} further introduces a specialized KG language via KGL token embedding augmentation. GraphGPT~\citep{tang2024graphgpt} proposes graph instruction tuning, tailored specifically for citation networks. In addition, InstructGLM~\citep{ye2023natural} and GRAG~\citep{hu2024grag} also attempt to combine graph encoding into LLMs. However, these methods often ignore the space inconsistency and struggle to effectively integrate structures of KGs, limiting LLMs' ability to fully understand and perform KGC tasks.






\section{Preliminary}

\noindent \textbf{Knowledge Graph Completion.}
Formally, a KG can be defined as $G=(\mathcal{E}, \mathcal{R},\mathcal{T})$, where $\mathcal{E}$ and $\mathcal{R}$ are the sets of entities and relations, respectively, and $\mathcal{T}=\{(h,r,t) \mid h,t \in \mathcal{E}, r \in \mathcal{R} \}$ represents the set of relational triples. Moreover, entities are often accompanied by rich textual information. In this paper, our work focuses on KGC, specifically addressing two tasks: (i) triple classification, assessing the correctness of a given triple $(h,r,t)$, and (ii) link prediction, identifying the most plausible tail entity $t$ for a given query $(h,r,?)$.

\noindent \textbf{Large Language Models.} 
LLMs have emerged as a powerful new paradigm for diverse tasks through prompting engineering. Let $\mathcal{M}$ denote the generative LLM, which takes a sequence $X=[x_1, ..., x_n]$ as the input prompt, and generates the response sequence $Y=[y_1, ..., y_m]$. Briefly, this process is represented as $Y = \mathcal{M}(X) $, where $X=[I; Q]$ consists of a task-specific instruction $I$ and an input query $Q$. This paper aims to develop a structure-aware LLM that excels in various KGC tasks by effectively integrating structural information.

\section{Methodology}

The architecture of SAT is illustrated in Figure~\ref{fig_framework}. Specifically, our model consists of two components: (1) hierarchical knowledge alignment and (2) structural instruction tuning. In the following sections, we provide a detailed explanation.


\subsection{Hierarchical Knowledge Alignment}\label{section_4_1}
To enhance LLMs' understanding of graph encoding, we consider local and global alignments.



\noindent \textbf{Local Knowledge Alignment.} To ensure that LLMs effectively understand the semantics of individual node representations, we align each entity with its corresponding textual description.





\noindent \textit{Node-Description Pairs Construction.}~For each entity $e$ in the KG $G$, we obtain its corresponding textual description $D_e$ by extracting the first paragraph from its linked Wikipedia page. For entities without linked pages, we use their names as alternative descriptions. In this way, we construct a set of node-description pairs $\mathcal{P}=\left\{\left(e, D_e\right) \mid e \in \mathcal{E}, D_e \in \mathcal{D} \right\}$, where $\mathcal{D}$ is the set of textual descriptions. 







\noindent \textit{Node-level Representation Alignment.}~We utilize a graph encoder, denoted as $\mathbf{GE}$, to encode the graph $G$ and obtain embeddings for entity nodes. Simultaneously, a text encoder, denoted as $\mathbf{TE}$, encodes the corresponding entity descriptions $\mathcal{D}$. In practice, we employ a graph transformer~\citep{yun2019graph} as the graph encoder and a vanilla transformer~\citep{vaswani2017attention} as the text encoder. The detailed encoding process is as follows\footnote{Note that the input to $\mathbf{GE}$ is the graph structure $G$, with node and edge features randomly initialized. Similarly, we apply the same approach for the global alignment in Eq.(\ref{equation_5}).},
\begin{equation}
\label{equation_1}
\setlength\abovedisplayskip{6pt}
\setlength\belowdisplayskip{6pt}
\mathbf{H}=\mathbf{GE}(G),
\end{equation}
\begin{equation}
\begin{aligned}
\setlength\abovedisplayskip{6pt}
\setlength\belowdisplayskip{6pt}
\mathbf{D} = \left\{ \mathbf{d}_e \mid \mathbf{d}_e = \mathbf{TE}(D_e),\; e \in \mathcal{E} \right\},
\end{aligned}
\end{equation}
\noindent where $\mathbf{H} \in \mathbb{R}^{N \times d}$ and $\mathbf{D} \in \mathbb{R}^{N \times d}$ represent the embeddings of nodes and descriptions, with $N$ denoting the number of nodes. We then apply L2 normalization to both $\mathbf{H}$ and $\mathbf{D}$. Finally, the node-level alignment is performed as follows,
\begin{equation}
\label{equation_3}
\setlength\abovedisplayskip{6pt}
\setlength\belowdisplayskip{3pt}
\boldsymbol{\Lambda}=(\mathbf{H} \, \mathbf{D}^{\top})\cdot \exp (\tau),
\end{equation}
\begin{equation}
\begin{aligned}
\setlength\abovedisplayskip{3pt}
\setlength\belowdisplayskip{6pt}
\mathcal{L}_{\mathrm{local}}= \frac{1}{2} \left( \operatorname{CE}(\boldsymbol{\Lambda}, \mathbf{Y}) + \operatorname{CE}(\boldsymbol{\Lambda}^{\top}, \mathbf{Y}) \right),
\end{aligned}
\end{equation}


\noindent where $\boldsymbol{\Lambda} \in \mathbb{R}^{N \times N}$ is the similarity matrix,  with each  $\boldsymbol{\Lambda}_{i,j}$ representing the similarity of the $i\text{-th}$ node embedding and the $j\text{-th}$ description embedding. The label $\mathbf{Y}=\mathbf{I}_N$ is the identity matrix, indicating the $i\text{-th}$ node embedding is closest to the $i\text{-th}$ description embedding. $\operatorname{CE(\cdot)}$ denotes the cross-entropy loss. Consequently, the loss $\mathcal{L}_{\mathrm{local}}$ encourages bidirectional alignment between nodes and their corresponding descriptions. In addition, the scalar $\tau \in \mathbb{R}$ is a temperature parameter.



\noindent \textbf{Global Knowledge Alignment.} To capture the global semantics conveyed by subgraphs, we focus further on global alignment that aligns subgraphs with their associated textual information.




\noindent \textit{Subgraph-Document Pairs Construction.}~Inspired by the powerful extraction abilities of LLMs, we employ GPT-4 to construct subgraph-document pairs. Initially, we sample a subset of the previously retrieved paragraphs to serve as input documents. For each document $D_S$, we leverage GPT-4 to extract triples\footnote{The extraction instruction is provided in Appendix~\ref{appendix_instruction_data}.}, which are then organized into the subgraph, denoted as $S$. Through this process, we construct a set of subgraph-document pairs, represented as $\mathcal{P^{\prime}}=\left\{\left(S, D_S\right) \mid S \in \mathcal{S}, D_S \in \mathcal{D}^{\prime}\right\}$, where $\mathcal{S}$ is the set of subgraphs and $\mathcal{D}^{\prime}$ is the set of corresponding documents.

\noindent \textit{Subgraph-level Representation Alignment.}~Similarly, we utilize the graph encoder $\mathbf{GE}$ and the text encoder $\mathbf{TE}$ to encode subgraphs $\mathcal{S}$ and textual documents $\mathcal{D}^{\prime}$  respectively. The subgraph embeddings $\mathbf{H}^{\prime}$ and document embeddings $\mathbf{D}^{\prime}$ are generated as follows,
\begin{equation}
\label{equation_5}
\mathbf{H}^{\prime} = \left\{ \mathbf{h}_S \;\middle|\; \mathbf{h}_S = \operatorname{Pool}(\mathbf{GE}(S)),\; S \in \mathcal{S} \right\},
\end{equation}
\begin{equation}
\mathbf{D}^{\prime} = \left\{ \mathbf{d}_S \mid \mathbf{d}_S = \mathbf{TE}(D_S),\; S \in \mathcal{S} \right\},
\end{equation}

\noindent where $\mathbf{H}^{\prime} \in \mathbb{R}^{M \times d}$, $\mathbf{D}^{\prime} \in \mathbb{R}^{M \times d}$, and $M$ is the number of subgraphs. $\operatorname{Pool}(\cdot)$ represents the mean pooling operator. Then, we calculate the similarity matrix $\boldsymbol{\Lambda}^{\prime}$ between $\mathbf{H}^{\prime}$ and $\mathbf{D}^{\prime}$ in the same way of Eq.(\ref{equation_3}), and apply the contrastive loss as follows,
\begin{equation}
\begin{aligned}
\setlength\belowdisplayskip{6pt}
\mathcal{L}_{\mathrm{global}}= \frac{1}{2} \left( \operatorname{CE}(\boldsymbol{\Lambda}^{\prime}, \mathbf{Y}) + \operatorname{CE}(\boldsymbol{\Lambda}^{\prime \top}, \mathbf{Y}) \right).
\end{aligned}
\end{equation}

The final loss is defined as a joint training objective of the local loss and the global loss:
\begin{equation}
\begin{aligned}
\mathcal{L_{\mathrm{HKA}}} = \mathcal{L}_{\mathrm{local}} +  \mathcal{L}_{\mathrm{global}}.
\end{aligned}
\end{equation}


Overall, we align graph embeddings with the natural language space. Note that the parameters of the local and global alignments are shared. 


\begin{figure}
    \centering
    \includegraphics[width=0.49\textwidth]{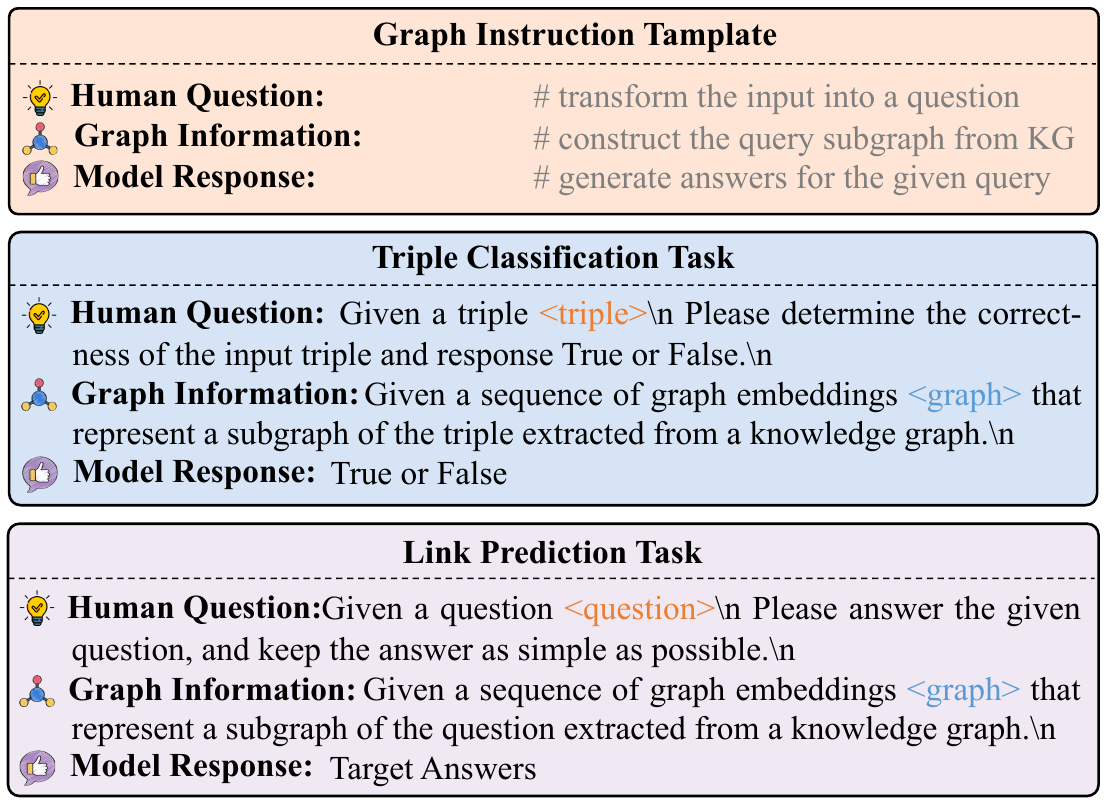}
    \caption{Illustration of graph instruction design. Graph instruction template unifies various KGC tasks.  
 } \label{fig_instruction}
\end{figure}

\subsection{Structural Instruction Tuning}\label{section_4_2}
To guide LLMs in structure-aware reasoning over KGs, we further introduce structural instruction tuning. Concretely, we design a unified graph instruction with a lightweight tuning strategy to support various KGC tasks.





\noindent {\textbf{Graph Instruction Design.}}
To unify KGC tasks, we formulate them as a generative question-answering problem, and design a graph instruction template comprising three parts: i) human question, ii) graph information, and iii) model response, as illustrated in Figure~\ref{fig_instruction}.
Specifically, given an input query (e.g., a triple), we first convert it into a human question $Q$. Next, to provide structural context, we construct the query subgraph $S_Q$ by extracting the $k$-hop neighborhoods around the anchor entities in the query from the KG.  The final input prompt is defined as $X=[I; Q; \textbf{EMB}(S_Q)]$, where $I$ represents the system instruction, and $\textbf{EMB}(\cdot)$ refers to the graph embeddings, obtained using the method introduced in Section~\ref{section_4_1}. This instruction template not only effectively integrates both natural language and structural information, but also flexibly unifies different KGC tasks.



\noindent {\textbf{Lightweight Tuning Strategy.}}
To optimize the fine-tuning process efficiently, we introduce a lightweight tuning strategy that incorporates a graph encoder and a knowledge adapter. The graph encoder is used to encode structural knowledge from the KG, while the adapter is designed to accommodate various KGC tasks. During training, we freeze the parameters of the LLM $\mathcal{M}$  and the graph encoder $\mathbf{GE}$, focusing solely on optimizing the adapter's parameters. Once trained, the LLM is expected to generalize across various tasks. In practice, the adapter can be implemented as a simple projection layer. Note that the graph encoder  $\mathbf{GE}$ is pre-trained as described in Section~\ref{section_4_1}.

In general, the fine-tuning objective is to maximize the model’s likelihood of generating the target response $Y$, conditioned on the prompt instruction $X$, as follows:
\begin{equation}
\setlength\abovedisplayskip{6pt}
\mathcal{L}_{\mathrm{SIT}}=\mathbb{E}_{(X, Y) \in (\mathcal{X}, \mathcal{Y})}\big[-\log P_{\mathcal{M}}(Y \mid X) \big ].
\end{equation}







\section{Experiments}

\subsection{Experimental Settings}
\noindent {\textbf{Datasets.}}
We evaluate SAT on two key KGC tasks: triple classification and link prediction, both of which are critical for completing incomplete KGs. Our experiments span four widely used benchmark datasets, categorized into two types: 1) small and hard (FB15k-237N, CoDeX-S) and  2) large and sparse (FB15k-237, YAGO3-10). The dataset statistics are presented in Table~\ref{tab_dataset}, and detailed descriptions can be found in Appendix \ref{appendix_datasets}.

\begin{table}[htbp]
\centering
\footnotesize
\setlength{\tabcolsep}{2.3pt}  
\caption{Statistics of the benchmark datasets.}
\label{tab_dataset}
\begin{tabular}{cccccc} 
\toprule
\textbf{Dataset}    & \textbf{\#Nodes}  & \textbf{\#Edges} & \textbf{\#Train} & \textbf{\#Valid} & \textbf{\#Test} \\ 
\midrule
FB15k-237N & 13,104    & 93      & 87,282   & 7,041    & 8,226 \\
CoDeX-S      & 2,034     & 42      & 32,888   & 1827    & 1,828 \\
\midrule
FB15k-237      & 14,505     & 237      & 272,115     & 17,535    & 20,466 \\

YAGO3-10      & 123,182     & 37      & 1,079,040   & 5,000    & 5,000 \\
\bottomrule
\end{tabular}
\end{table}

\noindent {\textbf{Baselines.}}~
We compare SAT with representative baselines grouping into two categories. The details of each baseline are described in Appendix \ref{appendix_baselines}.
\begin{itemize}[leftmargin=*, itemsep=0pt]
\item \textbf{Traditional KGC methods.} \textit{i) Embedding-based methods:} TransE~\citep{bordes2013translating},
ComplEx~\citep{trouillon2016complex}, 
ConvE~\citep{dettmers2018convolutional},
RotatE~\citep{sun2019rotate}, and
QuatE~\citep{sun2019rotate}.
\textit{ii) GNN-based methods:} 
CompGCN~\citep{vashishth2019composition}
and RED-GNN~\citep{zhang2022knowledge}. \textit{iii) Transformer-based methods:} PKGC~\citep{lv2022pre},
SimKGC~\citep{wang2022simkgc}, and
KG-S2S~\citep{chen2022knowledge}.

\item \textbf{LLM-enhanced KGC methods.} \textit{i) Text-based methods:} 
KG-LLaMA~\citep{yao2023exploring}, 
KERMIT~\citep{li2023kermit}, 
MPIKGC~\citep{xu2024multi}, 
GS-KGC~\citep{yang2024exploiting}, and 
KG-FIT~\citep{jiang2024kg}.
\textit{ii) Embedding-based methods:} CSProm-KG~\citep{chen2023dipping}, Structural-aware IT~\citep{zhang2024making}, and KoPA~\citep{zhang2024making}.
\end{itemize}

\noindent {\textbf{Evaluation Protocols.}}~
Following previous work, we evaluate triple classification using four standard metrics: accuracy, precision, recall, and F1-score, and assess link prediction with two metrics: Hits@1 and MRR.
Accuracy reflects the overall correctness, while  F1-score provides a balanced evaluation of precision and recall. Hits@1 measures the proportion of instances where the top-1 predicted answer is correct, and MRR calculates the average reciprocal rank of target answers.


\noindent {\textbf{Implementation Details.}}~For SAT, we use Llama2-Chat-7B~\citep{touvron2023llama} as the LLM backbone, fine-tuned on the training split of public datasets for 3 epochs. For each query, the top-3 answers are generated using a beam-search strategy. In the hierarchical knowledge alignment, we employ a graph transformer~\citep{yun2019graph} as the graph encoder, and a vanilla transformer~\citep{vaswani2017attention} as the text encoder. The pre-trained graph encoder and the learned graph embeddings are subsequently utilized in the structural instruction tuning module. During the tuning stage, the parameters of the LLM and the graph encoder are frozen, only the knowledge adapter is fine-tuned, implemented as a two-layer feed-forward neural network. Our model is implemented in Pytorch and trained on two Nvidia A800 GPUs. Detailed implementation settings are described in Appendix \ref{appendix_implementation}.




\begin{table*}
\centering
\footnotesize
\setlength{\tabcolsep}{4pt} 
\caption{ Experimental results on FB15k-237N and CoDeX-S datasets for triple classification. The best performances are highlighted in boldface, the second-best results are \underline{underlined}, and "*" indicates result reproduction.}
\label{tab_main_results}
\begin{tabular}{c|cccc|cccc} 
\toprule
\multirow{2}{*}{\textbf{Model}} & \multicolumn{4}{c|}{\textbf{FB15k-237N}}   & \multicolumn{4}{c}{\textbf{CoDeX-S}}  \\ 
\cline{2-9}
& \textbf{Accuracy}   & \textbf{Precision}  & \textbf{Recall}  & \textbf{F1-score}  & \textbf{Accuracy}   & \textbf{Precision}  & \textbf{Recall}  & \textbf{F1-score}  \\ 
\midrule
TransE~\citep{bordes2013translating}  & 0.697  & 0.708  & 0.671  & 0.689  & 0.721  & 0.719  & 0.724   & 0.722   \\
DistMult~\citep{yang2014embedding}   & 0.587  & 0.590  & 0.568  & 0.579  & 0.668  & 0.697  & 0.595  & 0.642    \\
ComplEx~\citep{trouillon2016complex}  & 0.657  & 0.665  & 0.634  & 0.649  & 0.676  & 0.678  & 0.671  & 0.675    \\
RotatE~\citep{sun2019rotate}  & 0.685  & 0.692  & 0.664  & 0.678  & 0.757  & 0.757  & 0.757  & 0.757     \\ 

KG-BERT~\citep{yao2019kg}  & 0.560  & 0.535  & \underline{0.976} & 0.678 & 0.773  & 0.710  & \textbf{0.924}  & 0.803    \\
PKGC~\citep{lv2022pre}$^*$  & \underline{0.796}  & 0.784  & 0.865  & 0.795   & 0.802  & 0.743  & \underline{0.923 } & 0.823   \\ 
\midrule
GPT-4~\citep{achiam2023gpt}$^*$   & 0.708  & \underline{0.817}  & 0.535  & 0.647  & 0.822  & \underline{0.825} & 0.703  & 0.798   \\
Vanilla IT~\citep{zhang2023making}  & 0.735  & 0.659  & 0.975  & 0.786  & 0.812  & 0.770  & 0.889  & 0.825   \\
KG-Alpaca~\citep{yao2023exploring} & 0.699  & 0.627  & \textbf{0.982}  & 0.766  & 0.803  & 0.794  & 0.817  & 0.805  \\ 
KG-LLaMA~\citep{yao2023exploring}   & 0.748  & 0.674  & 0.962  & 0.793  & 0.794  & 0.787  & 0.807  & 0.797  \\ 
Structural IT~\citep{zhang2024making}   & 0.764  & 0.696  & 0.940  & 0.799  & 0.813  & 0.771  & 0.884  & 0.826    \\
KoPA~\citep{zhang2024making}  & 0.777  & 0.708  & 0.941  & \underline{0.808} & \underline{0.827}  & 0.779  & 0.914  & \underline{0.841}   \\ 
\midrule
\textbf{SAT (ours)} & \textbf{0.827} & \textbf{0.823} & 0.852  & \textbf{0.831}   & \textbf{0.856} & \textbf{0.834}  & 0.893 & \textbf{0.865}  \\  


\bottomrule
\end{tabular}
\end{table*}

\subsection{Overall Comparison}

To evaluate our SAT, we conduct comprehensive experiments on two KGC tasks across four datasets.


\noindent \textbf{Triple Classification.}~The overall results are presented in Table~\ref{tab_main_results}. SAT significantly outperforms previous state-of-the-art methods, demonstrating the effectiveness of our model. When compared to PKGC, SAT achieves an average relative F1-score improvement of 4.5\% on FB15k-237N and 5.1\% on CoDeX-S,  highlighting its superiority over fully fine-tuned transformer-based models. In contrast with KoPA, our model shows average relative improvements of 2.8\% and 2.9\% on FB15k-237N and CoDeX-S, respectively.  These results emphasize the importance of aligning graph structure encoding with the natural language space to enhance the understanding of graph structures by LLMs. Additional experiment results on FB15k-237 and YAGO3-10 are provided in Appendix~\ref{appendix_tc_large}.

\noindent \textbf{Link Prediction.}~The results for link prediction are shown in Table~\ref{tab_lp_small_results} and Table~\ref{tab_lp_large_results}. In general, SAT substantially exceeds previous methods across all four datasets. Notably, SAT achieves significant relative improvements in Hit@1, surpassing the second-best model by 29.1\% on FB15k-237N and 29.8\% on CoDeX-S. Furthermore, compared to RED-GNN, our model demonstrates relative Hit@1 improvements of 13.1\% and 8.7\% on FB15k-237 and YAGO3-10, respectively. We attribute these substantial gains to two key factors: 1) our model effectively bridges the gap between graph structures and textual information through the hierarchical knowledge alignment, and 2) structural instruction tuning enables LLMs to comprehend the underlying structures within KGs, thus enhancing their graph reasoning abilities.

\begin{table}
\centering
\footnotesize
\setlength{\tabcolsep}{8pt}  
\caption{Results of link prediction on two small datasets.}
\label{tab_lp_small_results}
\begin{tabular}{l|cc|cc} 

\toprule
\multirow{2}{*}{\textbf{Model}} & \multicolumn{2}{c|}{\textbf{FB15k-237N }} & \multicolumn{2}{c}{\textbf{CoDeX-S }}  \\ 
\cline{2-5}
        & \textbf{Hit@1} & \textbf{MRR}     & \textbf{Hit@1} & \textbf{MRR}   \\ 
\midrule
TransE    & 0.152  & 0.255    & 0.219  & 0.354  \\
ConvE & 0.192  & 0.273    & 0.343  & 0.444  \\
TuckER     & 0.228  & 0.312    & 0.339  & 0.444  \\
CompGCN$^*$        & 0.231  & 0.316    & 0.315  & 0.395  \\
PKGC$^*$        & 0.261  & 0.332    & \underline{0.349}  & \underline{0.462}  \\
SimKGC$^*$        & 0.286  & 0.343    & 0.216  & 0.321  \\
KG-S2S  & 0.282  & 0.353    & -      & -      \\
HaSa$^*$  & \underline{0.299}  & \underline{0.374}    & 0.274      & 0.370      \\
GPT-4$^*$      & 0.241  & 0.264    & 0.196      & 0.232      \\
\midrule
\textbf{SAT  (ours)}    & \textbf{0.386} & \textbf{0.440}   & \textbf{0.453} & \textbf{0.481}         \\
\bottomrule
\end{tabular}
\end{table}

\begin{table}
\centering
\footnotesize
\setlength{\tabcolsep}{8pt}  
\caption{Results of link prediction on two large datasets.}
\label{tab_lp_large_results}
\begin{tabular}{l|cc|cc} 
\toprule
\multirow{2}{*}{\textbf{Model}} & \multicolumn{2}{c|}{\textbf{FB15k-237}} & \multicolumn{2}{c}{\textbf{YAGO3-10}}  \\ 
\cline{2-5}
   & \textbf{Hit@1} & \textbf{MRR}   & \textbf{Hit@1} & \textbf{MRR}  \\ 
\midrule
RotatE & 0.241  & 0.338    & 0.402  & 0.495 \\
CompGCN & 0.264  & 0.355  & 0.395  & 0.489 \\
RED-GNN & \underline{0.282} & \textbf{0.376}  & \underline{0.483}  & 0.559 \\
SimKGC$^*$ & 0.249  & 0.336   & 0.198 & 0.287      \\
HaSa$^*$ & 0.233  & 0.316   & 0.239 & 0.291     \\
CSProm-KG & 0.269  & 0.358  & 0.451  & 0.488 \\
KERMIT  & 0.266  & 0.359  & - & -   \\
MPIKGC & 0.267  & 0.360   & - & -     \\
KG-FIT & 0.275  & \underline{0.362}   & 0.474  & \underline{0.568} \\
\midrule
\textbf{SAT (ours)}  & \textbf{0.319} & 0.354 & \textbf{0.525} & \textbf{0.616}    \\
\bottomrule
\end{tabular}
\end{table}


\subsection{Ablation Study}
\begin{figure*}[htbp]
    \centering
	\subfigure[\scriptsize Effect of knowledge alignment]{
		\begin{minipage}[t]{0.24\linewidth}
        \centering
        
        \includegraphics[width=\textwidth]{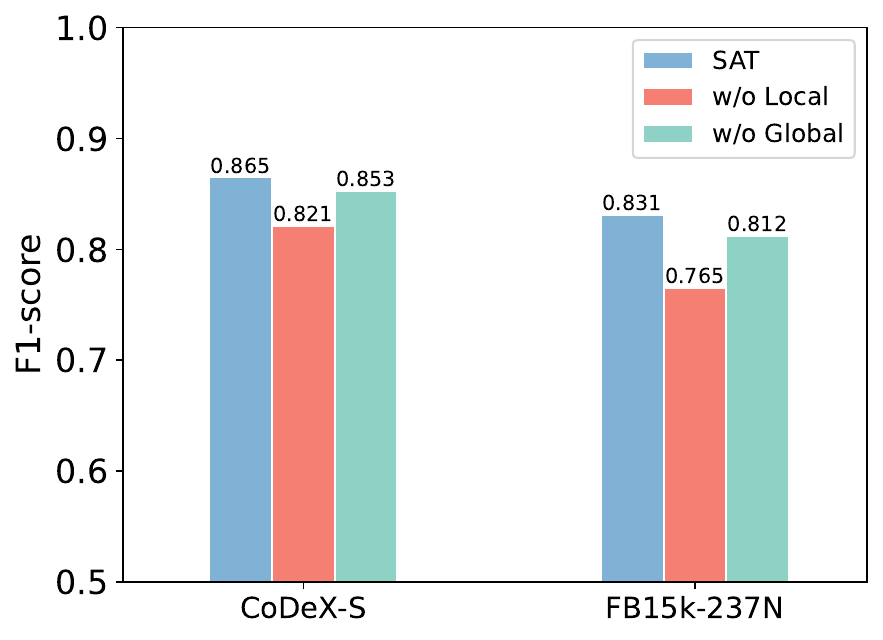}
		\end{minipage}%
	}%
	\subfigure[\scriptsize Effect of instruction design]{
		\begin{minipage}[t]{0.24\linewidth}
        \centering	
        \includegraphics[width=\textwidth]{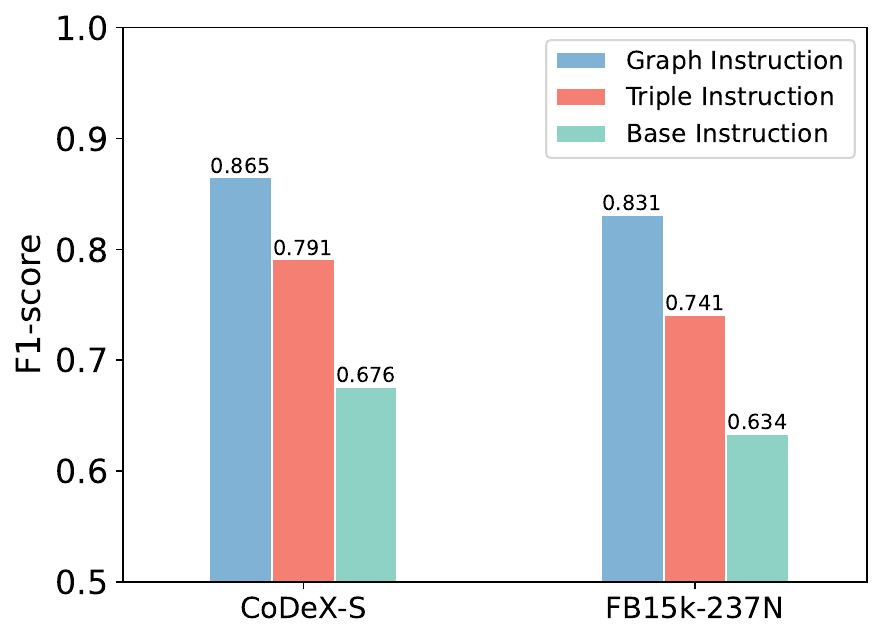}
		\end{minipage}
	}%
	\subfigure[\scriptsize Effect of external resources]{
		\begin{minipage}[t]{0.24\linewidth}
        \centering
        \includegraphics[width=\textwidth]{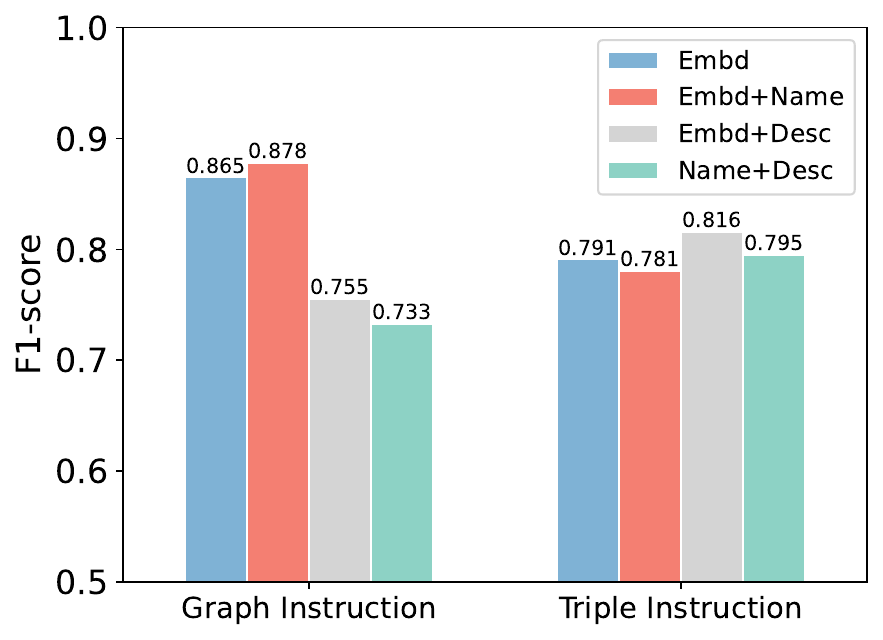}
		\end{minipage}%
	}%
	\subfigure[\scriptsize Effect of subgraph hops]{
		\begin{minipage}[t]{0.23\linewidth}
        \centering
		\includegraphics[width=\textwidth]{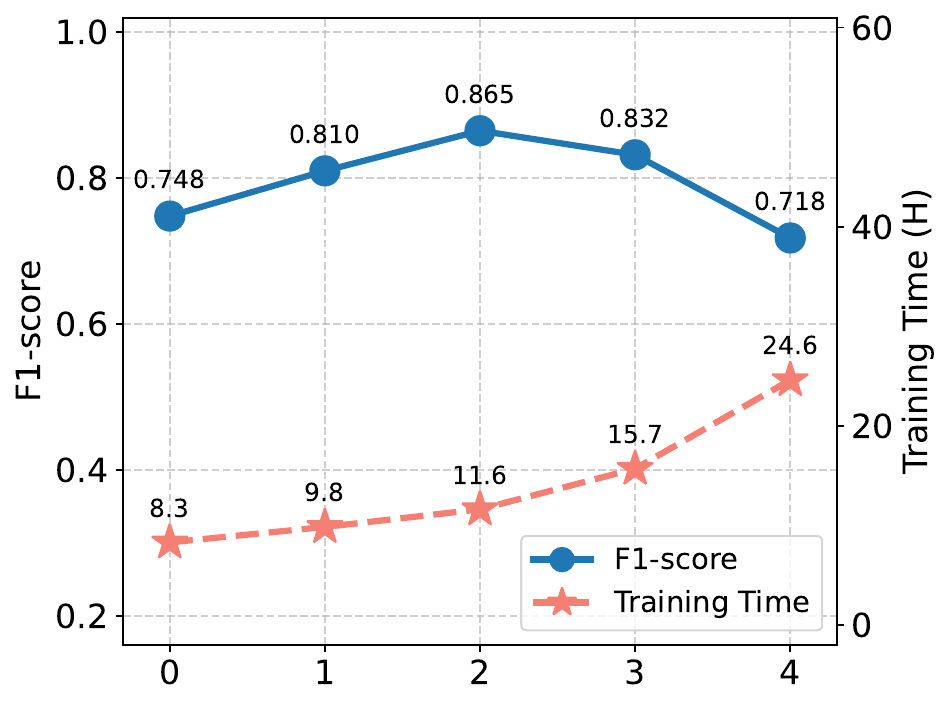}
		\end{minipage}
	}%
	\caption{The comparative experiments of SAT on triple classification. (a) Knowledge alignment removes local and global alignment, respectively. (b) Instruction design is based on subgraphs or individual triples. (c) External resources, such as names and descriptions, are incorporated. (d) Query subgraph is constructed with different hops.}
    \label{fig_ablation}
\end{figure*}


\noindent \textbf{Effect of Hierarchical Knowledge Alignment. }We conduct experiments to illustrate the necessity of aligning graph embeddings with the natural language space. As shown in Figure~\ref{fig_ablation}(a), the model without local alignment performs notably worse on FB15k-237N and CoDeX-S, which highlights the effectiveness of the node-level alignment. When global alignment is removed, the model struggles to understand global semantics expressed by subgraphs, resulting in a huge performance drop. This exhibits the critical importance of local and global alignment to achieve optimal results.



\noindent \textbf{Effect of Graph Instruction Design.}~Well-crafted instructions are crucial for enhancing the adaptability of LLMs. To this end, we propose three instruction modes: base instruction (without graph information), triple instruction (with a single triple), and graph instruction (with subgraph information).  As illustrated in Figure~\ref{fig_ablation}(b), the base instruction mode drops significantly, underscoring the importance of integrating graph information in KGC tasks. Furthermore, the graph instruction mode substantially outperforms the triple instruction mode, since query subgraphs provide richer contextual information than individual triples.
Detailed instructions are provided in Appendix~\ref{appendix_instruction_modes}.


\noindent \textbf{Effect of External Resources Fusion.}~ 
In addition to the graph information, we further enrich graph instruction by combining multiple external resources, such as graph embeddings, entity names, and text descriptions, as shown in Figure~\ref{fig_ablation}(c).  Interestingly, graph instructions augmented with entity names achieve the best results. However, when further supplemented with corresponding descriptions, the F1-score on CoDeX-S unexpectedly declines. We speculate that the additional descriptions may introduce extraneous information, potentially diluting the focus on critical entities. Meanwhile, this phenomenon also suggests that simply concatenating entity descriptions will disrupt semantic consistency and negatively impact model performance. Detailed descriptions of these instructions are illustrated in Appendix~\ref{appendix_instruction_resources}.

\begin{table}
\footnotesize
\centering
\setlength{\tabcolsep}{4pt}  
\caption{Robustness analysis of SAT under conditions of limited or noisy textual information.}
\label{tab_robustness}
\begin{tabular}{lccccc} 
\toprule
\multicolumn{1}{c}{\textbf{Description}} & \textbf{Linking} & \multicolumn{2}{c}{\textbf{FB15k-237N}} & \multicolumn{2}{c}{\textbf{CoDeX-S}}  \\ 
\cline{3-6}
\multicolumn{1}{c}{\textbf{Type}}  & \textbf{Noise}    & Hit@1 & MRR   & Hit@1 & MRR  \\ 
\midrule
Name   & 0\%  & 0.351 & 0.405  & 0.395 & 0.432   \\
Paragraph   & 0\%  & 0.386 & 0.440  & 0.453 & 0.481   \\
Paragraph   & 5\%  & 0.364 & 0.423  & 0.420 & 0.447   \\
Paragraph   & 10\%  & 0.328 & 0.397  & 0.372 & 0.413   \\
\bottomrule
\end{tabular}
\end{table}




\noindent \textbf{Effect of Query Subgraph Hops.}~~To evaluate the impact of subgraph scale, we construct subgraphs with different hops. As shown in Figure~\ref{fig_ablation}(d), the model's F1-score improves consistently as the number of hops increases. However, when the hop count exceeds 2, we observe a noticeable decline in performance, likely due to the inclusion of distant or noisy nodes. These findings indicate a trade-off between contextual richness and information noise. Therefore, we set the hop to 2 for subgraph construction in all subsequent experiments. In addition, we also investigate the impact of in-context learning in Appendix ~\ref{appendix_ict}. 






\subsection{Robustness Analysis}
In SAT, we rely on external textual information (e.g., Wikipedia), which may not always be available. To assess the robustness of our model under limited or noisy textual information, we present results in Table~\ref{tab_robustness}.
(1) \textit{Limited Textual Information}: We use only entity names as textual descriptions. Despite this reduction, SAT w/ name still achieves competitive inference performance, demonstrating strong generalization capabilities even in the absence of detailed contextual information. (2) \textit{Noisy Textual Information}: We simulate different levels of noise in the entity descriptions (e.g., random substitution of paragraphs) to evaluate model stability. The results show that SAT maintains reliable performance under moderate noise. We attribute this to the inherent graph structure, where contextual signals from neighboring entities could mitigate the impact of erroneous textual information.

\begin{table}
\footnotesize
\centering
\setlength{\tabcolsep}{4pt}  
\caption{Efficiency analysis on YAGO3-10 during both the model training and inference stages.}
\label{tab_efficiency}
\begin{tabular}{lccccc} 
\toprule
\multirow{2}{*}{\textbf{Methods}} & \multicolumn{1}{c}{\textbf{Language}} & \multicolumn{2}{c}{\textbf{\textbf{Training}}} & \multicolumn{2}{c}{\textbf{Inference}}  \\ 
\cline{3-6}
  & \multicolumn{1}{c}{\textbf{Model}}  & Params   & Time  & Time & \multicolumn{1}{l}{Hit@1}  \\ 
\hline
PKGC  & RoBERTa  & 353.1MB & 168H   & 50M & 0.156  \\
SimKGC  & BERT   & 106.2MB & 13H  & 10M & 0.198  \\
HaSa  & SBERT\footnotemark  & 218.0MB & 37H  & 4M  & 0.239  \\
Ours  & LLaMa2-7B   & 563.8MB & 70H  & 48M & 0.525  \\
\bottomrule
\end{tabular}
\end{table}
\footnotetext{SBERT (Sentence-BERT) is a modified version of the pretrained BERT network~\citep{reimers2019sentence}.}

\subsection{Efficiency Analysis}


To evaluate model efficiency, we conduct experiments on the larger-scale KG YAGO3-10. From Table~\ref{tab_efficiency}, we can observe that: (1) \textit{Computational Cost}: SAT is more efficient than PKGC, primarily due to the use of graph embeddings that effectively reduce prompt length. While SimKGC and HaSa exhibit lower computational time, our method achieves substantially better performance, highlighting a trade-off between efficiency and effectiveness. (2) \textit{Memory Usage}: SAT maintains the same order of magnitude (millions) as other methods. This is largely attributed to our lightweight fine-tuning strategy, in which only the adapter is updated. Overall, our method demonstrates acceptable time and space complexity. The specific parameters of our model are provided in Appendix~\ref{appendix_parameter}.

\subsection{Transferability Investigation}

\begin{figure}
	\centering
	\subfigure[\scriptsize Transferability on different LLMs]{
		\begin{minipage}[b]{0.47\linewidth}
			\centering
			\includegraphics[width=\textwidth]{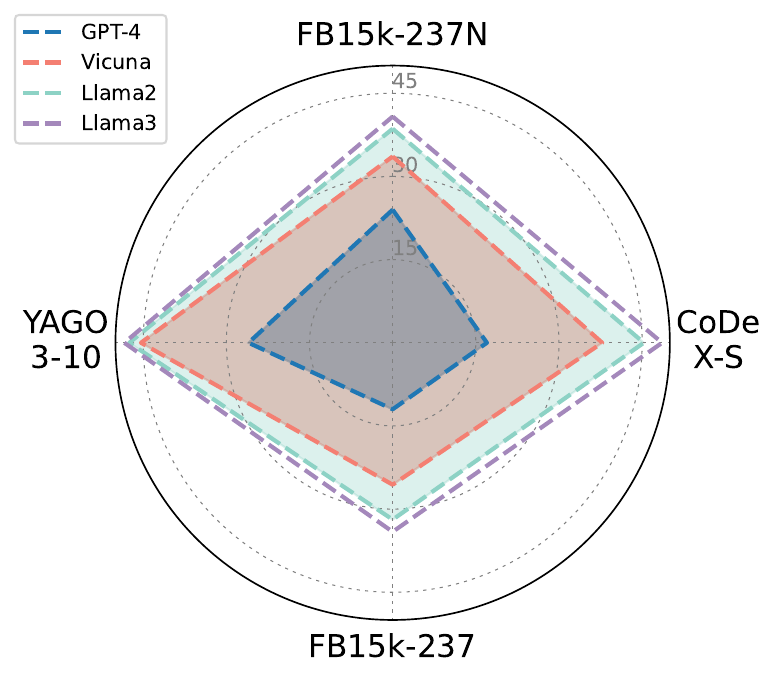}

		\end{minipage}%
	}%
    \hfill
	\subfigure[\scriptsize Transferability on multiple domains]{
    \centering
		\begin{minipage}[b]{0.49\linewidth}
			\centering
            \includegraphics[width=\textwidth]{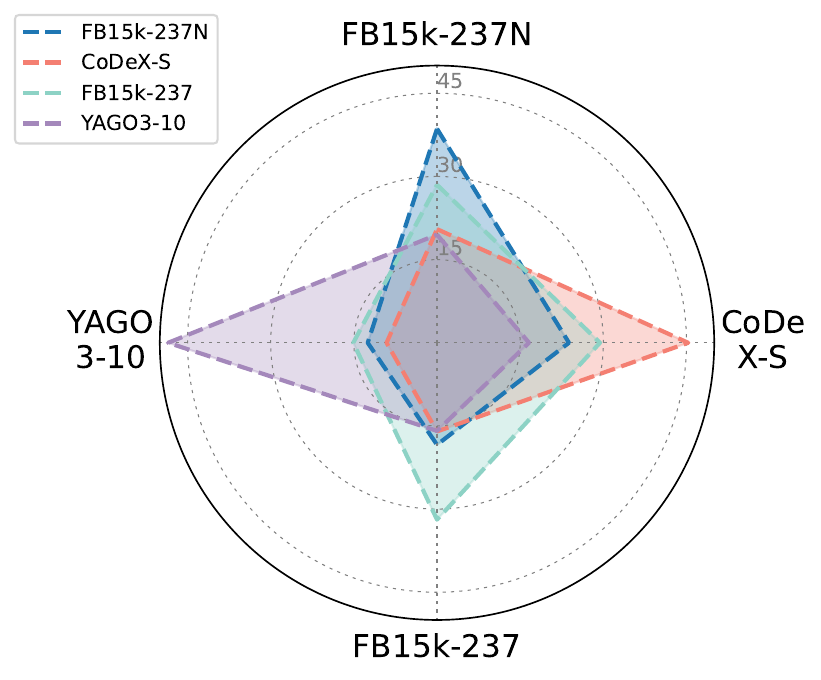}
		\end{minipage}
	}%
	\caption{Transferability study of our SAT on different LLMs and cross-domain datasets.}
    \label{fig_generalizability}
\end{figure}

To assess the transferability of SAT across different LLMs and cross-domain datasets, we carry out experiments in Figure \ref{fig_generalizability}.
(1) \textit{Transferability Across LLMs}. We evaluate link prediction on four datasets using representative LLMs, including Vicuna-v1.5-7B, Llama2-7B, and Llama3.1-8B. SAT consistently demonstrates remarkable performance across these models, underscoring its robustness and generalizability.
(2) \textit{Transferability Across Domains}. SAT exhibits strong graph reasoning capabilities when trained and tested on the same dataset. However, its performance degrades when applied to unseen datasets without further training. Notably, the model trained on FB15k-237N transfers better to FB15k-237 than to YAGO3-10, indicating that SAT achieves higher transferability between more closely related domains.

\begin{figure}
    \centering
    \includegraphics[width=0.38\textwidth]{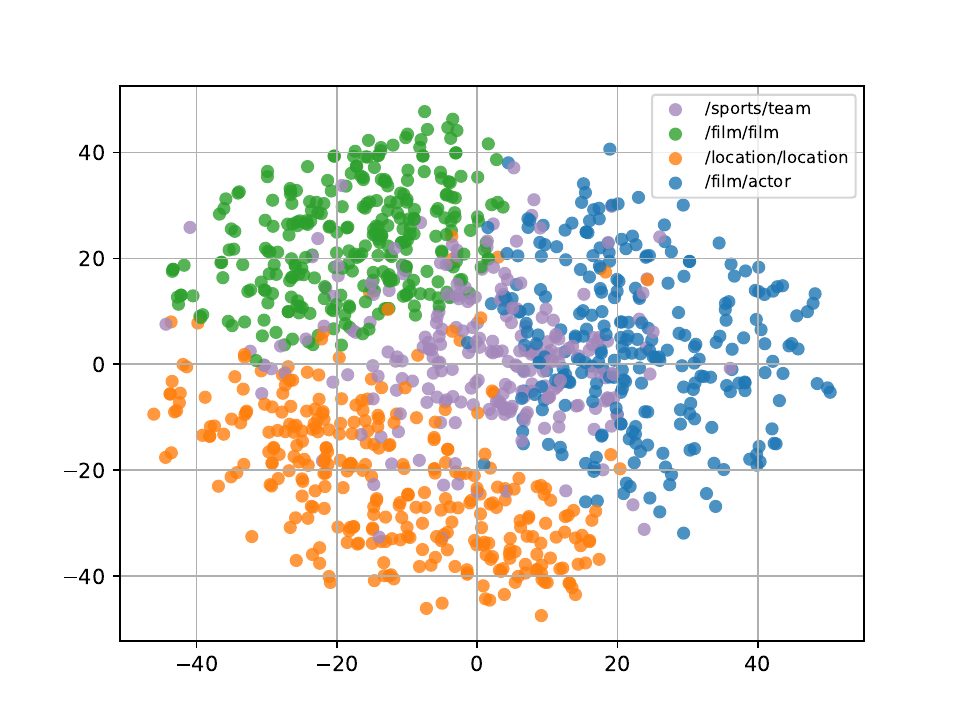}
    \caption{Visualization of graph embeddings.} \label{fig_visualization}
\end{figure}
\subsection{Representation Visualization}
We employ the t-SNE algorithm to visualize the learned graph embeddings, using a sample dataset comprising 1,000 entities across 4 classes from FB15k-237N. As shown in Figure~\ref{fig_visualization}, our method reveals more distinct class boundaries, particularly for \textit{/film/film}, \textit{/location/location}, and \textit{/film/actor}, demonstrating  the superiority of graph embeddings learned by our model and further enhancing LLMs' understanding of graph structure encoding. More visualizations are  included in the Appendix~\ref{appendix_visual}.



\begin{table}
\footnotesize
\centering
\caption{Case study of link prediction on FB15k-237N.}
\label{tab_casestudy}
\begin{tabularx}{\linewidth}{X} 
\toprule

\textbf{Task Instruction:}~You are an assistant specializing in large language models and knowledge graphs. Please follow the instructions carefully and provide your responses. \\ 

\hline
\multicolumn{1}{c}{{\cellcolor[rgb]{0.91,0.89,0.89}}\textbf{Text-based Instruction}}  \\

\textbf{Input:}~Given a question: \textbf{(Charles Dickens, influenced\_by, ?)}. Given a textual sequence of graph: (Charles Dickens, profession, Author), \textcolor{blue}{(Paul Auster, influenced\_by, Charles Dickens)}, \textcolor{blue}{(Leo Tolstoy, influenced\_by, Charles Dickens)}, (Paul Auster, profession, Author), (Paul Auster, influenced\_by, William Shakespeare) that represent a subgraph of the question extracted from a knowledge graph. Please answer the input question. \\
\textbf{Output (GPT-4):}~Paul Auster  and Leo Tolstoy ( \ding{55} )        \\

\hline
\multicolumn{1}{c}{{\cellcolor[rgb]{0.91,0.89,0.89}}\textbf{Embedding-based Instruction}}  \\

\textbf{Input:}~Given a question: \textbf{(Charles Dickens, influenced\_by, ?)}. Given a sequence of graph embeddings: \textcolor{red}{Emb(Charles Dickens)}, Emb(Paul Auster), Emb(Leo Tolstoy), \textcolor{red}{Emb(William Shakespeare)}, Emb(Author) that represent a subgraph of the question extracted from a knowledge graph. Please answer the input question. \\
\textbf{Output (Ours):}~William Shakespeare ( \ding{51} )     \\

\bottomrule
\end{tabularx}
\end{table}

\subsection{Case Study}
We present a case analysis on link prediction to further demonstrate the superiority of SAT. For the query (\textit{Charles Dickens}, \textit{influenced\_by}, ?), we compare two different methods: text-based instruction and embedding-based instruction. As illustrated in Table~\ref{tab_casestudy}, the embedding-based method effectively captures the implicit relationships between \textit{Charles Dickens} and \textit{William Shakespeare}, enabling accurate reasoning and predictions. Additionally, we test GPT-4 using the text-based instruction, but it incorrectly predicts two entities(i.e., \textit{Paul Auster} and  \textit{Leo Tolstoy}), likely due to misleading triples in the instruction, such as (\textit{Paul Auster}, \textit{influenced\_by}, \textit{Charles Dickens}) and (\textit{Leo Tolstoy}, \textit{influenced\_by}, \textit{Charles Dickens}). This indicates that GPT-4 is more sensitive to the specific wording of instructions compared to our model. 




\section{Conclusion}
In this paper, we introduce SAT, a novel framework designed to enhance LLMs for KGC tasks. Our model is comprised of two key components: hierarchical knowledge alignment and structural instruction tuning.  Extensive experiments on two KGC tasks across four datasets demonstrate the superiority of our method, achieving significant improvements over existing state-of-the-art baselines.

\section*{Limitation}
We analyze the limitations of our method and outline potential directions for future work. First, while our method effectively integrates graph embeddings into LLMs, we will explore more sophisticated strategies for combining textual and structural information to fully exploit the complementary strengths of both modalities. Second, although our SAT is a general framework adaptable to various KGC tasks, we plan to extend its applicability to a wider range of knowledge-intensive tasks and datasets, further enhancing its versatility.





\section*{Ethics Statement}
This work does not raise any direct ethical concerns. Our focus is on LLM-enhanced KGC tasks. All experiments are conducted on publicly available datasets, and the findings and conclusions of this paper are reported accurately and objectively.

\section*{Acknowledgments}
This work is supported by the National Key Research and Development Program of China (NO.2022YFB3102200), the National Natural Science Foundation of China (No.62402491), and the China Postdoctoral Science Foundation (No.2025M771524).


\bibliography{main}

\clearpage
\appendix

\section{Statistical information of datasets}\label{appendix_datasets}
In our experiments, we evaluate SAT on four popular benchmark datasets. The detailed descriptions of these datasets are provided below: 

\begin{itemize}[leftmargin=*, itemsep=0pt]
\item \textbf{FB15k-237N}~\citep{lv2022pre} is an improved version of FB15k-237 derived from Freebase, a large-scale KG containing factual knowledge.
\item \textbf{CoDeX-S}~\citep{safavi2020codex} is an encyclopedic KG extracted from Wikidata with three versions, among which we focus on CoDeX-S.
\item \textbf{FB15k-237}~\citep{toutanova2015observed}  is a subset of FB15k~\citep{bordes2013translating} with inverse relations removed to avoid data leakage.
\item \textbf{YAGO3-10}~\citep{suchanek2007yago} 
is a subset of YAGO, which is a large knowledge base derived from multiple external sources like Wikipedia.
\end{itemize}

\section{Detailed Descriptions of Baselines}\label{appendix_baselines}
The compared baselines can be divided into two categories: traditional KGC methods and LLM-enhanced KGC methods. For all compared methods, we provide a brief overview for reference.
\begin{itemize}[leftmargin=*, itemsep=0pt]
\item \textbf{Traditional KGC methods.} 
\textit{i) Embedding-based methods:} TransE~\citep{bordes2013translating} models relations as translations on entity embeddings. DistMult~\citep{yang2014embedding} introduces bilinear models for learning representations. ComplEx~\citep{trouillon2016complex} represents vectors in the complex space, while RotatE~\citep{sun2019rotate} extends this by modeling each relation as a rotation. Furthermore, several variants have been proposed to improve KGC performance, including QuatE~\citep{zhang2019quaternion}, HAKE~\citep{zhang2020learning}, HousE~\citep{li2022house}, and CompoundE~\citep{ge2023compounding}.
\textit{ii) GNN-based methods:}  GNN-based approaches, such as R-GCN~\citep{schlichtkrull2018modeling}, GraiL~\citep{teru2020inductive}, and CFAG~\citep{wang2022exploring}, leverage the inherent structural patterns of KGs. More recently, NBFNet~\citep{zhu2021neural} solves path formulation using the neural Bellman-Ford network. RED-GNN~\citep{zhang2022knowledge} introduces a novel relational directed graph for KG reasoning. \textit{iii) Transformer-based methods:} Most methods take advantage of PLMs by integrating textual information, such as KG-BERT~\citep{yao2019kg}, PKGC~\citep{lv2022pre}, and TAGREAL~\citep{jiang2023text}. Besides, SimKGC~\citep{wang2022simkgc}, HaSa~\citep{zhang2024hasa}, KRACL~\citep{tan2023kracl}, and MoCoSA~\citep{he2024mocosa} enhance PLMs through contrastive learning techniques.  


\item \textbf{LLM-enhanced KGC methods.} 
\textit{i) Text-based methods:} 
Recently, LLMs have been introduced to further improve KGC performance.
KG-LLaMA~\citep{yao2023exploring} first investigates KGC tasks using various LLMs. 
KICGPT~\citep{wei2024kicgpt} provides explicit instructions to guide the behavior of LLMs based on retrieved triples.
KERMIT~\citep{li2023kermit} employs LLMs for generating predictive descriptions, while MPIKGC~\citep{xu2024multi} prompts LLMs to generate auxiliary texts from various perspectives. Additionally, 
GS-KGC~\citep{yang2024exploiting} and KG-FIT~\citep{jiang2024kg} leverage LLMs for graph structure augmentation.
\textit{ii) Embedding-based methods:} 
CSProm-KG~\citep{chen2023dipping} investigates PLMs for KGC through structural soft prompts.
Structural-aware IT~\citep{zhang2024making} integrates structural knowledge into LLMs. 
KoPA~\citep{zhang2024making} incorporates structural encoding into LLMs through a prefix adapter. 
MKGL~\citep{guo2024mkgl2} further introduces a specialized KG language via KGL token embedding augmentation.
\end{itemize}


\section{Implementation Settings} \label{appendix_implementation}
In our implementation, we utilize Llama2-Chat-7B as the LLM backbone, fine-tuned on the training split of public datasets for 3 epochs. For hierarchical knowledge alignment, we employ a graph transformer as the graph encoder and a vanilla transformer as the text encoder. The local alignment is trained first, followed by the global alignment. During training, we set the learning rate to 1e-4, the dropout rate to 0.1, and use Adam optimizer. The dimensions for graph embedding and word embedding are set to 128, and the maximum text sequence length is limited to 256 tokens. In structural instruction tuning, the parameters of the pre-trained graph encoder and the LLM are frozen, while the knowledge adapter is fine-tuned exclusively during the instruction tuning stage. The training epoch is set to 3, the batch size is 2, and the learning rate is 2e-3, with a warmup ratio of 3e-2. Additionally, we sample 2-hop neighborhoods to build subgraphs, and the maximum input length for the LLM is capped at 2048 tokens. Our model is implemented in Pytorch and trained on two Nvidia A800 GPUs.

\section{Reproducibility  Explanation}
In the paper, we compare our method with representative baselines. To ensure a fair comparison, we report results from their original publications when available. For PKGC, CompGCN, SimKGC, and HaSa, we retrain the models using their official codes under the same experimental settings, as the original papers did not include results on some of the datasets used in our study. In addition, we evaluate the performance of GPT-4 (gpt-4-turbo) through OpenAI's publicAPI. 

\section{Experimental Results}
\subsection{Triple Classification on Large Datasets}\label{appendix_tc_large}
Table~\ref{tab_tc_large_results} presents the results for triple classification on FB15k-237 and YAGO3-10. As shown, SAT achieves relative F1 improvements of 1.3\% and 2.4\% over PKGC, respectively. Additionally, when compared to GPT-4, our model also demonstrates 14.9\% and 5.8\% relative improvements. These findings further emphasize the effectiveness of our model across diverse datasets.

\begin{table}[ht]
\centering
\footnotesize
\setlength{\tabcolsep}{9pt}  
\caption{Results of triple classification on large datasets.}
\label{tab_tc_large_results}
\begin{tabular}{l|cc|cc} 
\toprule
\multirow{2}{*}{\textbf{Model}} & \multicolumn{2}{c|}{\textbf{FB15k-237}} & \multicolumn{2}{c}{\textbf{YAGO3-10}}   \\ 
\cline{2-5}
  & \textbf{Acc} & \textbf{F1}    & \textbf{Acc} & \textbf{F1}  \\ 

\midrule
PKGC$^*$    & 0.823     & 0.846  & 0.807   & 0.831    \\ 
GPT-4$^*$    & 0.726     & 0.615 & 0.781   & 0.643    \\ 
\textbf{SAT (ours)}    & 0.834   & 0.822 & 0.826   & 0.795    \\

\bottomrule
\end{tabular}
\end{table}


\subsection{Model Parameter Supplement}\label{appendix_parameter}
We provide a detailed supplement of model parameters in Table~\ref{tab_parameter}. As shown, compared to the knowledge alignment phrase (GNN+PLM), the instruction tuning phrase (LLM) results in a substantial increase in training parameters (525M, 8.0\%) and time (23H). This underscores that tuning the LLM is a significant computational bottleneck.

\subsection{Global Alignment Data Statistics}
The statistics for global alignment data are summarized in Table~\ref{tab_global_data}. It comprises 12,000 pairs, with an average processing time of 2.41 seconds per document. Note that FB15k-237N and FB15-237 share the same pairs. On average, each document contains approximately 490 tokens, and the corresponding subgraph includes around 12 triples.

\begin{table}[ht]
\footnotesize
\centering
\setlength{\tabcolsep}{1.5pt}  
\caption{Model parameters of SAT during training stage.}
\label{tab_parameter}
\begin{tabular}{lccc} 
\toprule
\textbf{Model}      & \textbf{Time} & \textbf{Params(\%)} & \textbf{Params(\#)}  \\ 
\midrule
Ours (Tune GNN+PLM) & 5H   & 1.0\%  & 65,360,896       \\
Ours (Tune LLM)     & 23H  & 8.0\%  & 525,893,632      \\
Ours (Tune All)     & 28H  & 8.9\%  & 591,254,528      \\
Ours (Pretrain All) & OOM  & 100\%  & 6,609,987,584    \\
\bottomrule
\end{tabular}
\end{table}

\begin{table}[ht]
\footnotesize
\centering
\setlength{\tabcolsep}{3.2pt} 
\caption{Statistical analysis of global alignment data. }
\label{tab_global_data}
\begin{tabular}{cccccc} 
\toprule
\textbf{Datasets} & \textbf{Pairs(\#)} & \textbf{Tokens(\#)} & \textbf{Triples(\#)}  & \textbf{Time(H)}   \\ 
\midrule
CoDeX-S  & 1,000    & 518   & 12 & 0:42:37 \\
FB15k-237 & 6,000   & 481   & 11  & 3:55:12 \\
YAGO3-10 & 5,000    & 495  & 12 & 3:23:20 \\
\bottomrule
\end{tabular}
\end{table}

\subsection{More Representation Visualizations}\label{appendix_visual}
To visualize the aligned representation space, we further incorporate more visualizations, using a new sample dataset comprising about 500 entities across 3 classes from FB15k-237N.

\begin{figure}[ht]
    \centering
    \includegraphics[width=0.42\textwidth]{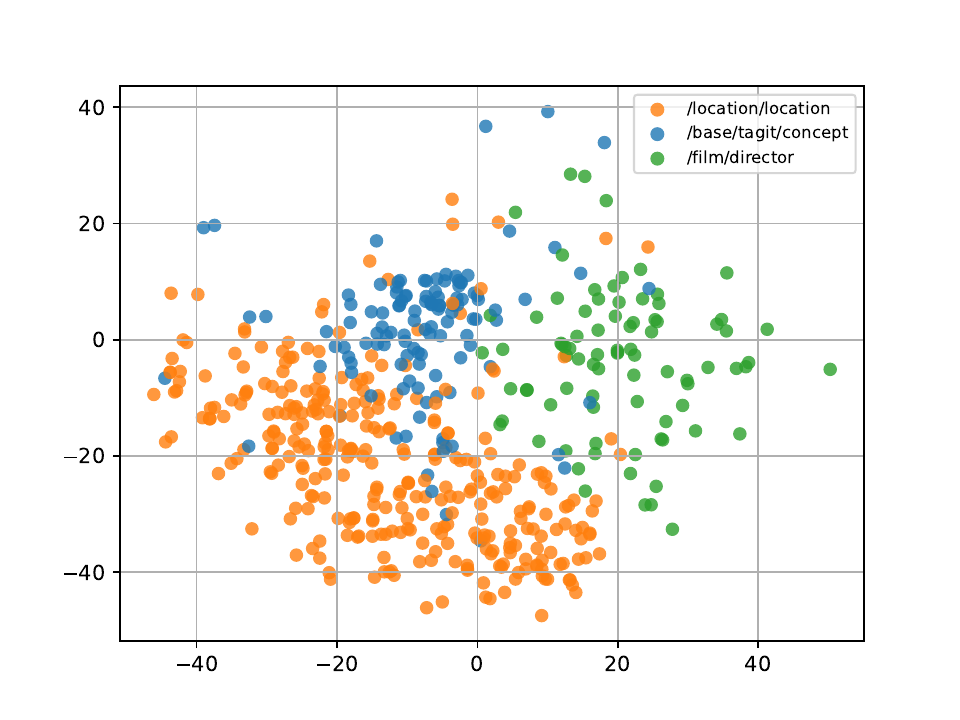}
    \caption{Visualization of graph embeddings.} \label{fig_visualization_more}
\end{figure}


\section{In-Context Learning}
\label{appendix_ict}
Recently, in-context learning (ICL) has emerged as a powerful technique that incorporates task examples directly into the prompts. However, the effectiveness of ICL is significantly dependent on the selection of relevant examples. Unlike previous methods that rely primarily on the textual semantics of the query, we propose the structural ICL, which retrieves structure-related examples. Specifically, we build a candidate pool $\mathcal{C}=\{C_1, C_2, ...\}$, where each candidate is represented as $\mathcal{C}_i=(Q_i, S_i)$. We then select the top-$k$ relevant examples based on the similarity $\varphi$ between the current query subgraph $S_q$ and each candidate subgraph $S_i$, as follows,
\begin{equation}
\mathcal{K}_{\text {top-k }}=\underset{\left\{C_i\right\}_{i=1}^k}{\operatorname{argmax}} \;  \sum_{i=1}^k \varphi\left(\boldsymbol{h}_{S_q}, \boldsymbol{h}_{S_i}\right).
\end{equation}

\begin{figure}
	\centering
	\subfigure[\scriptsize Effect of structural ICL]{
		\begin{minipage}[t]{0.5\linewidth}
			\centering
			\includegraphics[width=\textwidth]{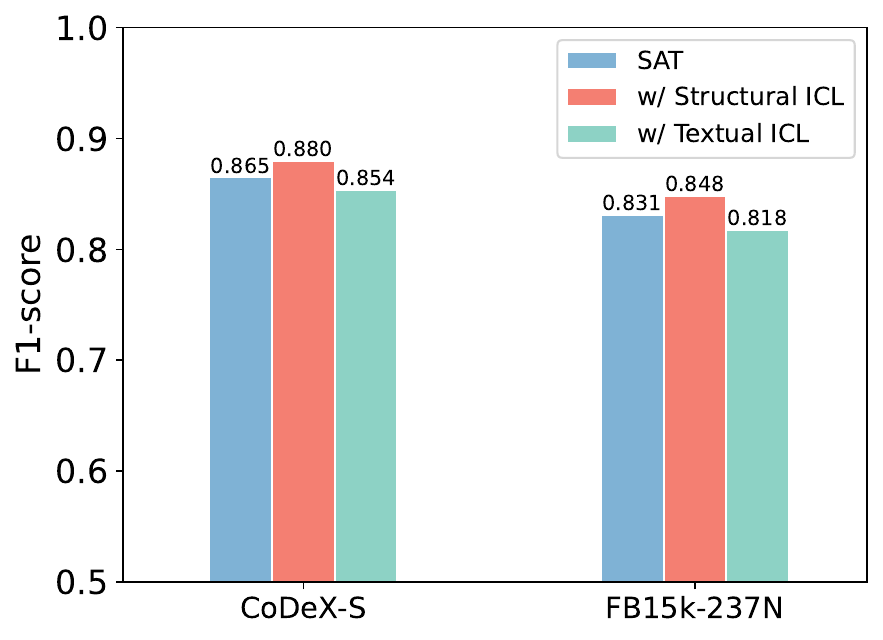}

		\end{minipage}%
	}%
	\subfigure[\scriptsize  Effect of example numbers ]{
    \centering
		\begin{minipage}[t]{0.5\linewidth}
			\centering
            \includegraphics[width=\textwidth]{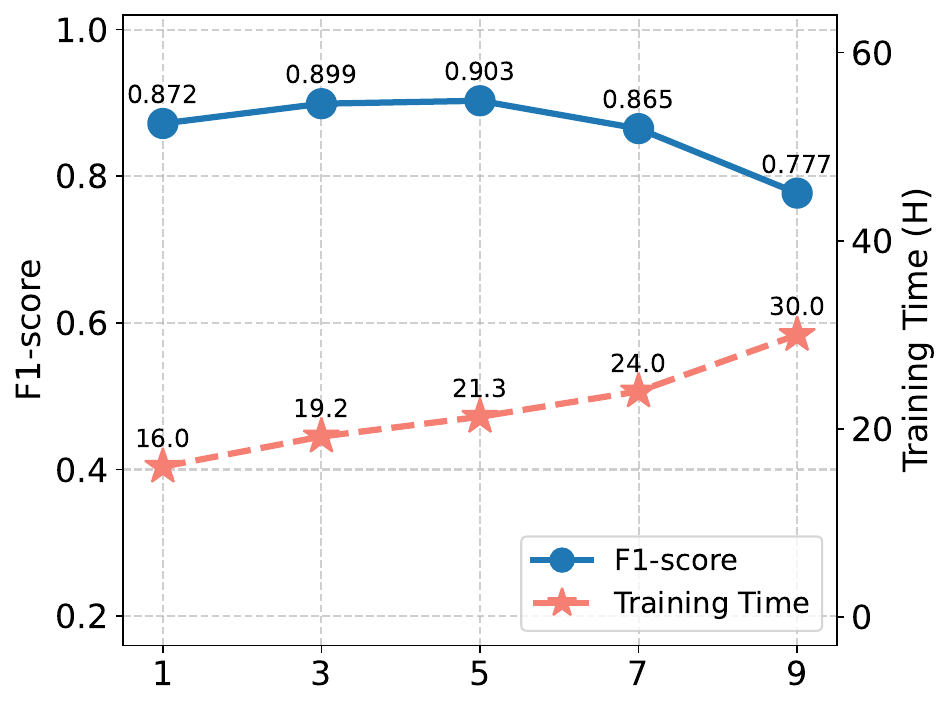}
		\end{minipage}
	}%
	\caption{Additional experiments on ICL. }
    \label{fig_appendix_ict}
\end{figure}

\noindent \textbf{Effect of Structural ICL.}~ To assess the impact of structural ICL, we implement experiments with different retrieval mechanisms. As shown in Figure~\ref{fig_appendix_ict}(a), we observe a noticeable improvement when structural ICL is introduced.  Moreover, structural ICL outperforms textual ICL, which we attribute to the fact that textual similarity is insufficient for capturing graph structures. In contrast, structural ICL can effectively retrieve structure-related examples. Note that these experiments are conducted using 3-shot examples.


\noindent \textbf{Effect of Example Numbers.}~~As known, the effectiveness of ICL is significantly influenced by the number of retrieved examples. To determine the optimal number of examples, we perform various experiments on CoDeX-S, as shown in Figure~\ref{fig_appendix_ict}(b). Our findings reveal that while the training time increases linearly with the number of instances, the performance does not exhibit a corresponding upward trend. To balance effectiveness and efficiency, we set the optimal ICL number to 3.

\section{Instruction Templates}
\subsection{Instruction for Data Construction} \label{appendix_instruction_data}
We design an instruction to prompt GPT-4 to extract subgraphs from the provided documents for the construction of global alignment data.

\subsection{Graph Instruction for KGC tasks}
We present instruction templates for two key KGC tasks: triple classification and link prediction. 

\subsection{Graph Instruction with Different Graph Information}\label{appendix_instruction_modes}
We design three modes for graph instruction: base instruction (without graph information), triple instruction (with a single triple), and graph instruction (with subgraph information).
Templates for triple classification are provided, with templates for link prediction developed using the same way.

\subsection{Graph Instruction with Multiple External Resources}\label{appendix_instruction_resources}
In addition to the graph structure,  we further enrich graph instruction by incorporating external resources, such as entity names and text descriptions.
Again, specific templates for triple classification are provided, while templates for link prediction are adapted from this framework. 

\begin{figure}[ht]
\centering
\begin{tcolorbox}[ colback=white,colframe=black,left=3pt,right=3pt,
title=Instruction templates for global data construction, fonttitle=\footnotesize]
\footnotesize
\textbf{Task Instruction:}~Given a document, please extrapolate as many relationships as you can from the document and generate triples like (source, relation, target).

\bigskip
{\bfseries \textcolor{blue!70!black}{Example}}\\
{\textbf{Input:}~Steven Paul Jobs was an American businessman, inventor, and investor best known for co-founding the technology giant~Apple Inc.~Jobs was also the founder of~NeXT~and chairman and majority shareholder of~Pixar. He was a pioneer of the~personal computer revolution~of the 1970s and 1980s, along with his early business partner and fellow Apple co-founder~Steve Wozniak.} \\
\textbf{Output:}~(Steven Paul Jobs, nationality, American); (Steven Paul Jobs, occupation, businessman); (Steven Paul Jobs, occupation, inventor); (Steven Paul Jobs, co-founder of, Apple Inc.); (Steven Paul Jobs, founder of, NeXT); (Steven Paul Jobs, chairman of, Pixar); (Steven Paul Jobs, business partner, Steve Wozniak);

\bigskip
\textbf{\textbf{Input:}}~Original Document \\
\textbf{Output:}~Generated Triples 
\end{tcolorbox}
\end{figure}

\begin{figure}[ht]
\centering
\begin{tcolorbox}[ colback=white,colframe=black,left=5pt,right=5pt,title=Instruction templates for two KGC tasks, fonttitle=\footnotesize]
\footnotesize
\textbf{Task Instruction:}~You are an assistant specializing in large language models and knowledge graphs. Please follow the instructions carefully and provide your responses.

\bigskip
{\bfseries \textcolor{green!50!black}{Triple Classification}}\\
\textbf{Input:}~Given a triple \{\textit{triple}\} that consists of a head entity, a relation, and a tail entity. Given a sequence of graph embeddings \{\textit{graph}\} that represent a subgraph of the triple extracted from a knowledge graph. Please determine the correctness of the input triple and response True or False.\\ 
\textbf{Output:}~True or False

\bigskip
{\bfseries \textcolor{violet}{Link Prediction}}\\
\textbf{Input:}~Given a question:\{\textit{question}\} that represents a natural language question. Given a sequence of graph embeddings \{\textit{graph}\} that represent a subgraph of the question extracted from a knowledge graph. Please answer the input question, and keep the answer as simple as possible.\\ 
\textbf{Output:}~Target Answers
\end{tcolorbox}
\end{figure}

\begin{figure}[ht]
\centering
\begin{tcolorbox}[ colback=white,colframe=black,left=5pt,right=5pt,title=Instruction templates for graph information, fonttitle=\footnotesize]
\footnotesize
\textbf{Task Instruction:}~You are an assistant specializing in large language models and knowledge graphs. Please follow the instructions carefully and provide your responses.

\bigskip
{\bfseries \textcolor{blue!70!black}{Base Instruction}}\\
\textbf{Input:~}Given a triple \{\textit{triple}\} that consists of a head entity, a relation, and a tail entity. Please determine the correctness of the input triple and response True or False.\\ 
\textbf{Output:}~True or False

\bigskip
{\bfseries \textcolor{green!50!black}{Triple Instruction}}\\
\textbf{Input:~}Given a triple \{\textit{triple}\} that consists of a head entity, a relation, and a tail entity. Given a sequence of graph embeddings \{\textit{graph}\} that represent the anchor entities of the triple. Please determine the correctness of the input triple and response True or False.\\ 
\textbf{Output:}~True or False

\bigskip
{\bfseries \textcolor{violet}{Graph Instruction}}\\
\textbf{Input:}~Given a triple \{\textit{triple}\} that consists of a head entity, a relation, and a tail entity. Given a sequence of graph embeddings \{\textit{graph}\} that represent a subgraph of the triple extracted from a knowledge graph. Please determine the correctness of the input triple and response True or False.\\ 
\textbf{Output:}~True or False
\end{tcolorbox}
\end{figure}


\begin{figure}[ht]
\centering
\begin{tcolorbox}[ colback=white,colframe=black,left=5pt,right=5pt,title=Instruction templates for external sources, fonttitle=\footnotesize]
\footnotesize
\textbf{Task Instruction:}~You are an assistant specializing in large language models and knowledge graphs. Please follow the instructions carefully and provide your responses.

\bigskip
{\bfseries \textcolor{blue!70!black}{Entity Names}}\\
\textbf{Input:~}Given a triple \{\textit{triple}\} that consists of a head entity, a relation, and a tail entity. Given a sequence of graph embeddings \{\textit{graph}\} that represent a subgraph of the triple extracted from a knowledge graph. Each graph node contains an entity name. Here is a list of entity names: \{\textit{name}\}. Please determine the correctness of the input triple and response True or False.\\ 
\textbf{Output:}~True or False

\bigskip
{\bfseries \textcolor{green!50!black}{Text Descriptions}}\\
\textbf{Input:~}Given a triple \{\textit{triple}\} that consists of a head entity, a relation, and a tail entity. Given a sequence of graph embeddings \{\textit{graph}\} that represent a subgraph of the triple extracted from a knowledge graph. Each graph node contains an entity description. Here is a list of entity textual descriptions: \{\textit{description}\}. Please determine the correctness of the input triple and response True or False.\\ 
\textbf{Output:}~True or False

\bigskip
{\bfseries \textcolor{violet}{Entity Names + Text Descriptions}}\\
\textbf{Input:~}Given a triple \{\textit{triple}\} that consists of a head entity, a relation, and a tail entity. Given a subgraph of the triple extracted from a knowledge graph. Each graph node contains an entity name and its textual description information. Here is a list of entity information: \{\textit{name}\} and \{\textit{description}\}. Please determine the correctness of the input triple and response True or False.\\ 
\textbf{Output:}~True or False
\end{tcolorbox}
\end{figure}

\end{document}